\documentclass[runningheads]{llncs}

\usepackage{graphicx}
\usepackage{booktabs}
\usepackage{multirow}
\usepackage{amsmath,amssymb}
\usepackage{xcolor}
\usepackage{url}
\usepackage[expansion=false]{microtype}

\usepackage[hidelinks,breaklinks=true]{hyperref}
\usepackage[capitalize]{cleveref}
\graphicspath{{figures/}}
\newcommand{\method}{GAMF-Net}

\hypersetup{
    pdftitle={GAMF-Net: Grouped Availability-Aware Multimodal Fusion for Smartphone Skin Lesion Classification},
    pdfauthor={Yue Zhuo, Jingjia Yan, Fangyu Liu, Lingjun Liu, Zongjun Chen, Zhisheng Luo, Zhonghua Xie},
    pdfsubject={Multimodal skin lesion classification},
    pdfkeywords={skin lesion classification, multimodal fusion, clinical metadata, availability-aware learning}
}

\AddToHook{cmd/normalsize/after}{%
    \setlength{\abovedisplayskip}{5pt plus 2pt minus 2pt}%
    \setlength{\belowdisplayskip}{5pt plus 2pt minus 2pt}%
    \setlength{\abovedisplayshortskip}{3pt plus 2pt}%
    \setlength{\belowdisplayshortskip}{4pt plus 2pt minus 2pt}%
}
\AtBeginDocument{%
    \setlength{\emergencystretch}{2em}%
    \setlength{\abovedisplayskip}{3pt plus 2pt minus 1pt}%
    \setlength{\belowdisplayskip}{3pt plus 2pt minus 1pt}%
    \setlength{\abovedisplayshortskip}{2pt plus 1pt}%
    \setlength{\belowdisplayshortskip}{3pt plus 1pt minus 1pt}%
    \setlength{\textfloatsep}{8pt plus 2pt minus 2pt}%
    \setlength{\floatsep}{8pt plus 2pt minus 2pt}%
    \setlength{\intextsep}{10pt plus 2pt minus 2pt}%
}

\begin{document}

\title{
GAMF-Net: Grouped Availability-Aware Multimodal Fusion
for Smartphone Skin Lesion Classification
}

\titlerunning{
GAMF-Net for Smartphone Skin Lesion Classification
}

\author{
Yue Zhuo \and
Jingjia Yan \and
Fangyu Liu \and
Lingjun Liu \and
Zongjun Chen \and
Zhisheng Luo \and
Zhonghua Xie\thanks{Corresponding author: Zhonghua Xie (email: eexie@qq.com).}
}
\authorrunning{Y. Zhuo et al.}
\institute{Huizhou University, Huizhou 516007, Guangdong, China\\
\email{eexie@qq.com}}

\maketitle

\begin{abstract}
Smartphone clinical images provide a low-cost data source for computer-aided skin lesion classification in resource-constrained settings, but images alone cannot fully capture patients' medical histories and lesion symptoms. To address the lack of structured organization among heterogeneous clinical variables, the ambiguity between observed and imputed values, and the risk that clinical information may interfere with visual evidence, we propose \method, a grouped availability-aware multimodal fusion network. Ten clinical variables are divided into three semantic groups---demographics, history-related risk, and lesion presentation---and each group jointly encodes field values and field-level availability masks. The image branch produces the base classification logits, while each clinical group generates image-conditioned class-level residual evidence. Only available groups participate in availability-normalized equal-weight fusion; when all clinical information is unavailable, the model reduces exactly to the image-only pathway. We evaluate the method on PAD-UFES-20 under a patient-disjoint five-fold protocol with separate training, validation, and test partitions. \method{} achieves an Accuracy of $74.17\pm2.85\%$, a balanced accuracy (BACC) of $70.11\pm2.11\%$, a Macro-F1 of $68.85\pm0.85\%$, and a Macro-AUROC of $93.42\pm1.20\%$, outperforming all image-only and multimodal baselines reimplemented under a unified protocol. Compared with the strongest baseline, MetaBlock, the corresponding improvements are 2.37, 2.59, 4.14, and 1.02 percentage points, respectively. Ablation results show that clinical information substantially improves class-balanced performance, constrained equal-weight fusion is more effective than dynamic fusion, and the lesion-presentation group provides the most informative clinical complement.
\end{abstract}

\section{Introduction}

Deep convolutional neural networks have demonstrated strong performance for automated skin lesion recognition on curated clinical and dermoscopic image collections~\cite{esteva2017dermatologist,han2018clinical}. Nevertheless, most widely used public benchmarks, including HAM10000, are dominated by dermoscopic images acquired under comparatively controlled conditions~\cite{tschandl2018ham10000}. Smartphone clinical photographs are easier to acquire in primary-care and resource-constrained settings, but they exhibit greater variation in illumination, background, camera distance, device type, and lesion scale. PAD-UFES-20 was introduced to support research in this less controlled setting by providing smartphone images together with patient- and lesion-level clinical variables~\cite{pacheco2020pad}.

Clinical decision-making relies on information that is not always visually observable, such as age, previous cancer history, anatomical location, growth, bleeding, pain, and elevation. Multimodal learning provides a general framework for combining complementary sources of evidence~\cite{baltrusaitis2019multimodal}. In skin lesion analysis, prior studies have explored image--metadata concatenation, metadata-conditioned feature modulation, and attention-based interaction. Yap et al.\ combined multiple imaging modalities with patient metadata~\cite{yap2018multimodal}; Li et al.\ used metadata-derived coefficients to reweight visual feature channels~\cite{li2020metanet}; and MetaBlock introduced an attention-based metadata-processing block for image feature enhancement~\cite{pacheco2021metablock}. Subsequent studies have explored more expressive interactions using multimodal Transformers, MDFNet, and the cross-attention-based MMF-Net~\cite{cai2022transformer,chen2022mdfnet,ou2022mmfnet}. Although these approaches confirm the value of clinical context, they generally encode heterogeneous variables as a single vector or allow metadata to directly perturb visual features.

A further difficulty is incomplete clinical information: standard imputation produces fixed-dimensional inputs but does not indicate whether a value was observed or imputed. MetaBlock-SE represents available metadata with sentence embeddings and evaluates robustness under simulated and real missingness~\cite{bouzon2025metablockse}, while TIP develops image--tabular pre-training for heterogeneous and incomplete tabular data~\cite{du2024tip}. More generally, explicit missingness masks can provide information beyond imputed values~\cite{che2018grud}. These methods motivate availability-aware modeling, but preserving an independent visual decision path while incorporating semantically grouped clinical evidence remains insufficiently explored.

To address these limitations, we propose \method, which organizes ten clinical variables into groups representing demographic information, history-related risk, and lesion presentation. Each group is encoded together with field-level availability masks. The image branch first produces a complete set of six-class logits, and each clinical group then generates image-conditioned class-level residual evidence; only available groups participate in normalized equal-weight fusion, so clinical information supplements rather than replaces the image prediction.

The main contributions are as follows:
\begin{itemize}
    \item We introduce medically motivated semantic grouping and field-availability-aware encoding to jointly model clinical values and their observation states.
    \item We develop an image-anchored class-level residual fusion mechanism in which clinical groups provide complementary corrections, while the model reduces exactly to the image-only pathway when all clinical groups are unavailable.
    \item We conduct a unified evaluation on PAD-UFES-20 using patient- and lesion-disjoint five-fold splits, including image-only models, multimodal baselines, and fusion/input and clinical-group ablation comparisons.
\end{itemize}

\section{Related Work}

\subsection{Image-Based Skin Lesion Classification}

Deep learning has substantially advanced automated skin lesion
classification. Esteva et al.\ demonstrated dermatologist-level
discrimination on selected tasks using convolutional networks trained
on large clinical image collections~\cite{esteva2017dermatologist}.
Han et al.\ further investigated benign and malignant cutaneous tumor
classification using standard clinical photographs rather than
dermoscopic images alone~\cite{han2018clinical}. For dermoscopic
analysis, HAM10000 provides 10,015 multi-source images and remains
a major benchmark for multiclass lesion recognition
~\cite{tschandl2018ham10000}. These studies establish the effectiveness
of image-based learning, but they do not directly address the greater
acquisition variability and richer clinical context associated with
smartphone images.

PAD-UFES-20 contains 2,298 smartphone clinical images from
1,373 patients and 1,641 lesions across six classes, together with
up to 21 clinical variables~\cite{pacheco2020pad}. The dataset is
strongly class-imbalanced and may contain multiple images of the
same lesion, so patient- and lesion-disjoint evaluation is necessary
to prevent information leakage and obtain a realistic estimate of
generalization performance.

\subsection{Fusion of Images and Clinical Metadata}

Multimodal learning includes early fusion, late fusion,
representation alignment, and modality-conditioned interaction
~\cite{baltrusaitis2019multimodal}. In skin lesion classification,
direct concatenation is a common baseline, but it does not explicitly
distinguish the roles of visual and clinical evidence. Yap et al.\
showed that clinical metadata can complement multiple skin-imaging
modalities~\cite{yap2018multimodal}. MetaNet uses a metadata network
to generate channel-wise modulation coefficients for visual features
~\cite{li2020metanet}, whereas MetaBlock employs an attention-based
metadata-processing block to enhance image representations
~\cite{pacheco2021metablock}.

More expressive fusion mechanisms have subsequently been proposed.
Cai et al.\ introduced a multimodal Transformer for image--metadata
fusion~\cite{cai2022transformer}; MDFNet combines skin images and
clinical information through a dedicated multimodal diagnosis
framework~\cite{chen2022mdfnet}; and MMF-Net employs intra-modality
self-attention and inter-modality cross-attention for bidirectional
interaction~\cite{ou2022mmfnet}. These methods strengthen cross-modal
communication, but their additional flexibility can complicate
optimization on small and imbalanced datasets. In contrast,
\method{} preserves an explicit image decision pathway and introduces
clinical information as group-specific residual evidence at the
logit level.

\subsection{Incomplete Image--Tabular Data}

Incomplete clinical information poses a further challenge because
measurements may not be collected, verified, or retained during data
curation; conventional imputation preserves a fixed-dimensional input
but does not indicate whether each value was observed or imputed. GRU-D demonstrated that explicit masks and missingness
patterns can improve healthcare prediction beyond conventional
imputation~\cite{che2018grud}. Although GRU-D addresses multivariate
time series rather than image--metadata fusion, its treatment of
observation states supports explicit availability indicators.

MetaBlock-SE extends MetaBlock using sentence embeddings and
evaluates robustness to incomplete metadata on PAD-UFES-20 and an
extended clinical dataset~\cite{bouzon2025metablockse}. TIP addresses
a broader image--tabular setting through masked tabular
reconstruction, image--tabular matching, and contrastive pre-training
~\cite{du2024tip}. Rather than reconstructing or semantically
embedding missing values, \method{} directly exposes field
availability to each clinical group encoder and excludes completely
unavailable groups from residual fusion.

\subsection{Recent Advances in Medical and Consumer Imaging}

Medical image computing has increasingly relied on generative and
learned priors for reconstruction and synthesis~\cite{chen2026image,11297154,11098517},
and visual enhancement has been extended to robot-assisted
surgery~\cite{li2026rga-net}. Low-level restoration studies address
document contamination removal~\cite{chen2026docunfold}, facial
shadow removal~\cite{zeng2026dynroutenet,11462271}, cinematic filter
emulation~\cite{11460866}, and underwater~\cite{chen2026aquaclarity}
and infrared imagery~\cite{11523523}. On the multimodal side, recent
work has examined robustness to attention sinks and positional decay
in large vision-language models~\cite{zhang2026tokenpenalty},
redundancy-aware cross-modal emotion
recognition~\cite{11586066}, manifold-guided interaction in consumer
devices~\cite{xu2026mage}, and persona-driven speech
synthesis~\cite{li2026atrie}. Further studies cover counseling
alignment~\cite{11543262}, 4D Gaussian splatting
compression~\cite{11609274}, and compositional concept
extraction~\cite{DBLP:conf/cvpr/ShiTHJCPWP26}. None models the
field-level availability of heterogeneous clinical variables, the
focus of \method.

\section{Method}

\begin{figure}[!t]
    \centering
    \vspace{-2mm}

    \includegraphics[
        width=0.90\linewidth,
        trim=12mm 18mm 2mm 4mm,
        clip
    ]{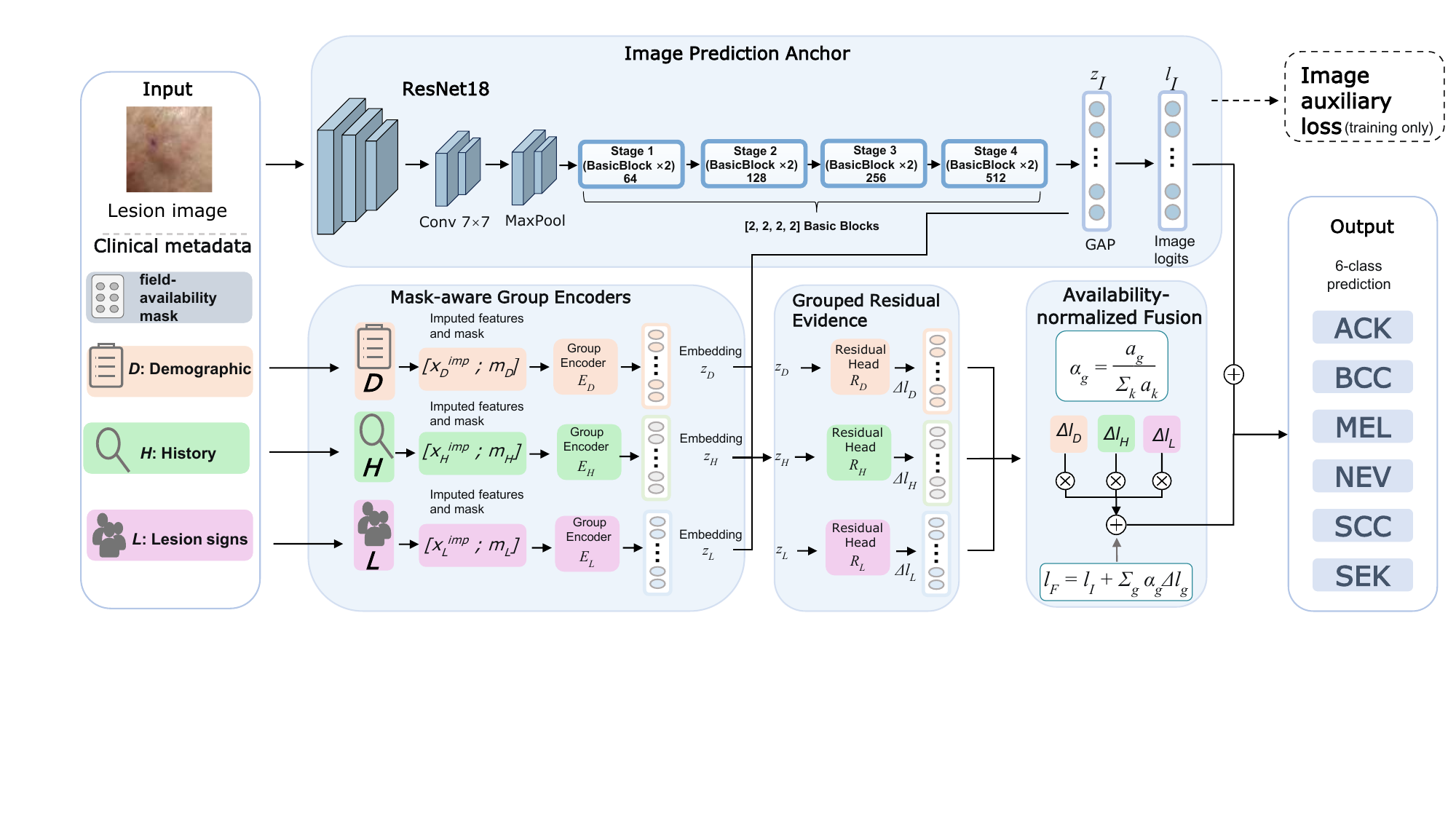}

    \vspace{-8mm}
    \caption{
    Overview of \method.
    Overview of \method: image-anchored prediction
    with availability-aware grouped clinical residuals.
    }
    \label{fig:overview}

    \vspace{-3mm}
\end{figure}

\subsection{Problem Formulation and Clinical Grouping}

Let the set of clinical semantic groups be $\mathcal{G}=\{D,H,L\}$, corresponding to demographics, history-related risk, and lesion presentation, respectively. Given a sample
\begin{equation}
\left(\mathbf{x}^{I},\{\mathbf{x}^{g},\mathbf{m}^{g}\}_{g\in\mathcal{G}},y\right),
\end{equation}
where $\mathbf{x}^{I}$ denotes a smartphone clinical image, $\mathbf{x}^{g}$ contains the clinical fields in group $g$, $\mathbf{m}^{g}$ is the corresponding field-level availability mask, and $y$ is one of the six lesion labels, the model learns
\begin{equation}
p_{\theta}\!\left(y\mid \mathbf{x}^{I},\{\mathbf{x}^{g},\mathbf{m}^{g}\}_{g\in\mathcal{G}}\right).
\end{equation}

We use a compact set of ten clinical variables:
\begin{align}
D &= \{\texttt{age}\},\\
H &= \{\texttt{skin\_cancer\_history},\texttt{cancer\_history}\},\\
L &= \{\texttt{region},\texttt{itch},\texttt{grew},\texttt{hurt},
\texttt{changed},\texttt{bleed},\texttt{elevation}\}.
\end{align}

Imputation and standardization parameters for continuous variables are estimated only from the training subset of the current fold, missing categorical values are assigned to a dedicated category before one-hot encoding, and the validation and test subsets are transformed only with the preprocessor fitted on the training subset. The fields \texttt{patient\_id}, \texttt{lesion\_id}, and \texttt{img\_id} are used only for data organization, \texttt{diagnostic} is the prediction target, and \texttt{biopsed} is closely associated with diagnostic confirmation; none is used as model input.

\subsection{Field-Availability-Aware Group Encoding}

Suppose group $g$ contains $q_g$ original fields. Its field mask, field-availability ratio, and group-level availability indicator are defined as
\begin{equation}
\mathbf{m}_g\in\{0,1\}^{q_g},\qquad
r_g=\frac{1}{q_g}\sum_{j=1}^{q_g}m_{g,j},\qquad
a_g=\mathbb{I}(r_g>0).
\end{equation}

The preprocessed clinical features are concatenated with the field-level mask:
\begin{equation}
\widetilde{\mathbf{x}}_g=[\mathbf{x}^{\mathrm{imp}}_g;\mathbf{m}_g].
\end{equation}

In the ten-field configuration, the preprocessed feature dimensions of the D, H, and L groups are 1, 6, and 32 and their mask dimensions are 1, 2, and 7, giving encoder input dimensions of 2, 8, and 39.

The three group encoders share the same architecture but have independent parameters:
\begin{equation}
\widehat{\mathbf{z}}_g
=
\operatorname{GELU}\!\left(
\operatorname{LN}\!\left(
\mathbf{W}^{(2)}_g
\operatorname{Dropout}_{0.2}\!\left[
\operatorname{GELU}\!\left(
\operatorname{LN}(
\mathbf{W}^{(1)}_g\widetilde{\mathbf{x}}_g+\mathbf{b}^{(1)}_g)
\right)\right]
+\mathbf{b}^{(2)}_g
\right)\right),
\end{equation}
where the first layer maps the input to 256 dimensions, and the second layer maps it to 128 dimensions. The final group representation is
\begin{equation}
\mathbf{z}_g=a_g\widehat{\mathbf{z}}_g\in\mathbb{R}^{128}.
\end{equation}

The field mask enables the encoder to distinguish genuine observations from imputed placeholders, following the principle that missingness indicators may contain predictive information beyond imputed values~\cite{che2018grud,du2024tip}; multiplication by $a_g$ further ensures that a completely unavailable group does not produce an effective representation.

\subsection{Image Anchor and Grouped Residual Evidence}

The input image is resized to $3\times224\times224$ and encoded by a ResNet18 pretrained on ImageNet-1K~\cite{he2016resnet,deng2009imagenet}:
\begin{equation}
\mathbf{z}_{I}=E_I(\mathbf{x}^{I})\in\mathbb{R}^{512},
\qquad
\mathbf{l}_{I}
=
\mathbf{W}_{I}\operatorname{Dropout}_{0.3}(\mathbf{z}_{I})+\mathbf{b}_{I}
\in\mathbb{R}^{6}.
\end{equation}

The vector $\mathbf{l}_{I}$ serves as the image prediction anchor for the final multimodal decision.

For each clinical group, the residual head takes the image representation, the group representation, and the field-availability ratio as input:
\begin{equation}
\mathbf{u}_g=[\mathbf{z}_{I};\mathbf{z}_g;r_g]\in\mathbb{R}^{641},
\end{equation}
\begin{equation}
\Delta\mathbf{l}_g
=
a_g R_g(\mathbf{u}_g)
\in\mathbb{R}^{6}.
\end{equation}

The residual mapping $R_g$ is a $641\!\rightarrow\!256\!\rightarrow\!6$ MLP with LayerNorm, GELU, and Dropout in the hidden layer. Its final linear layer is initialized with zero weights and zero bias, so that $\Delta\mathbf{l}_g=\mathbf{0}$ at the beginning of training. The clinical branch is therefore initialized to preserve the image-only prediction and gradually learns image-conditioned class corrections.

\subsection{Availability-Normalized Fusion and Training Objective}

Let $s=\sum_{k\in\mathcal{G}}a_k$. The fusion weight of group $g$ is
\begin{equation}
\alpha_g=
\begin{cases}
a_g/s,&s>0,\\
0,&s=0.
\end{cases}
\end{equation}
The final logits are
\begin{equation}
\mathbf{l}_F
=
\mathbf{l}_I+
\sum_{g\in\mathcal{G}}\alpha_g\Delta\mathbf{l}_g.
\end{equation}

When all clinical groups are unavailable, $\mathbf{l}_F=\mathbf{l}_I$; unlike a learned dynamic gate, the fusion assigns weights only by group availability, limiting unnecessary model flexibility in the small-data regime.

The training objective is
\begin{equation}
\mathcal{L}
=
\operatorname{CE}_{\mathbf{w}}(\mathbf{l}_F,y)
+
\lambda_I
\operatorname{CE}_{\mathbf{w}}(\mathbf{l}_I,y),
\qquad
\lambda_I=0.3,
\end{equation}
where the class weights are computed from the class frequencies of the training subset in the current fold. The image auxiliary loss maintains the standalone discriminative capacity of the image anchor.

\section{Experimental Setup}

\subsection{Dataset and Evaluation Protocol}

PAD-UFES-20 contains 2,298 smartphone images from 1,373 patients and 1,641 lesions~\cite{pacheco2020pad}, distributed as 730, 845, 52, 244, 192, and 235 images in the ACK, BCC, MEL, NEV, SCC, and SEK classes, respectively, resulting in substantial class imbalance.

\begin{figure}[ht]
    \centering
    \includegraphics[width=0.97\linewidth]{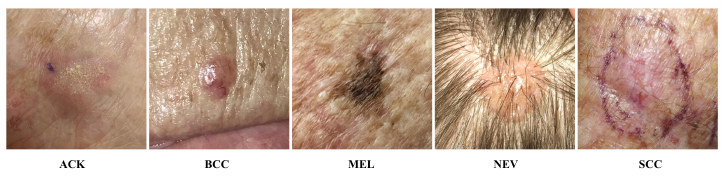}
    \caption{Representative smartphone clinical images of the five skin lesion classes in PAD-UFES-20.}
    \label{fig:representative}
\end{figure}

We use a fixed patient-disjoint five-fold cross-validation protocol: in each fold, three folds are used for training, one for validation, and one for testing, containing 1,371--1,391, 449--475, and 449--475 images, respectively, with no patient or lesion shared across subsets. The preprocessor is fitted only on the training subset, the best model is selected based on validation BACC, and the selected checkpoint is evaluated once on the test subset. Final results are reported as the mean and sample standard deviation across the five test folds.

\subsection{Compared Methods and Implementation Details}

\begin{sloppypar}
We compare \method{} against Image-only ResNet18, Direct Concat, MetaNet~\cite{li2020metanet}, Dynamic Feature Modulation, MetaBlock-SE~\cite{bouzon2025metablockse}, and MetaBlock~\cite{pacheco2021metablock}; Direct Concat is a conventional early-fusion baseline, and Dynamic Feature Modulation a unified metadata-conditioned feature-modulation baseline. All reimplemented models use the same patient-level splits, class order, field-exclusion rules, preprocessing procedures, and validation-based model-selection criterion.
\end{sloppypar}

\method{} uses a ResNet18 pretrained on ImageNet with an input resolution of $224\times224$ and a batch size of 32. AdamW~\cite{loshchilov2019adamw} is used as the optimizer, with learning rates of $1\times10^{-5}$ for the backbone and $1\times10^{-4}$ for the newly added modules and a weight decay of $1\times10^{-4}$. We use a OneCycle learning-rate schedule motivated by super-convergence training~\cite{smith2018superconvergence}, train for at most 100 epochs with early stopping (patience 15), set the random seed to 2026, and enable automatic mixed precision. Experiments use Python 3.10, PyTorch 2.7.1, CUDA 12.6, and an NVIDIA Tesla V100 32GB GPU.

\subsection{Evaluation Metrics}

We report Accuracy, BACC, Macro-F1, and Macro-AUROC. BACC is appropriate for imbalanced classification because it averages class-wise sensitivity~\cite{brodersen2010balanced}. It is defined as follows:
\begin{equation}
\mathrm{BACC}=\frac{1}{C}\sum_{c=1}^{C}\mathrm{Recall}_c.
\end{equation}

Because the dataset is class-imbalanced, validation BACC is used for model selection. Class-wise recall and AUROC are also reported to analyze minority-class discrimination across decision thresholds~\cite{fawcett2006roc}.

\section{Results and Discussion}

\subsection{Main Results}

\begin{table}[t]
\centering
\caption{
Patient-level five-fold test results
(mean $\pm$ sample standard deviation, \%).
Bold values indicate the highest mean under the unified protocol.
}
\label{tab:main}
\setlength{\tabcolsep}{4.0pt}
\resizebox{\linewidth}{!}{%
\begin{tabular}{lcccc}
\toprule
Model & Accuracy & BACC & Macro-F1 & Macro-AUROC\\
\midrule
Image-only ResNet18
& $64.93\pm2.67$
& $61.53\pm3.11$
& $58.06\pm2.69$
& $88.69\pm1.20$\\
Direct Concat
& $62.88\pm4.10$
& $61.65\pm4.42$
& $55.89\pm3.49$
& $87.84\pm1.65$\\
MetaNet
& $68.84\pm4.97$
& $64.32\pm2.22$
& $62.03\pm3.13$
& $90.71\pm1.33$\\
Dynamic Feature Modulation
& $68.07\pm3.49$
& $62.89\pm2.22$
& $60.94\pm2.39$
& $89.75\pm1.38$\\
MetaBlock-SE
& $66.56\pm1.49$
& $62.12\pm3.86$
& $58.86\pm3.00$
& $88.92\pm1.62$\\
MetaBlock
& $71.80\pm1.74$
& $67.52\pm2.64$
& $64.71\pm3.06$
& $92.40\pm0.95$\\
\textbf{\method{}}
& $\mathbf{74.17\pm2.85}$
& $\mathbf{70.11\pm2.11}$
& $\mathbf{68.85\pm0.85}$
& $\mathbf{93.42\pm1.20}$\\
\bottomrule
\end{tabular}
}
\end{table}

As shown in \cref{tab:main}, \method{} achieves the highest mean
on all four metrics among the models evaluated under the unified
protocol. Compared with MetaBlock, \method{} improves Accuracy,
BACC, Macro-F1, and Macro-AUROC by 2.37, 2.59, 4.14, and
1.02 percentage points, respectively. Relative to the image-only
ResNet18, BACC and Macro-F1 increase by 8.58 and 10.79 points,
demonstrating the complementary value of structured clinical
information under class imbalance.

Direct Concat performs below the image-only model on Accuracy,
Macro-F1, and Macro-AUROC, indicating that simply appending
clinical variables does not guarantee effective multimodal learning.
MetaNet~\cite{li2020metanet} and MetaBlock~\cite{pacheco2021metablock}
improve performance through metadata-conditioned feature modeling,
whereas \method{} separates the visual prediction from group-specific
clinical residual evidence.

\subsection{Structural Ablations and Paired Analysis}

\begin{figure}[t]
    \centering
    \includegraphics[width=0.96\linewidth]{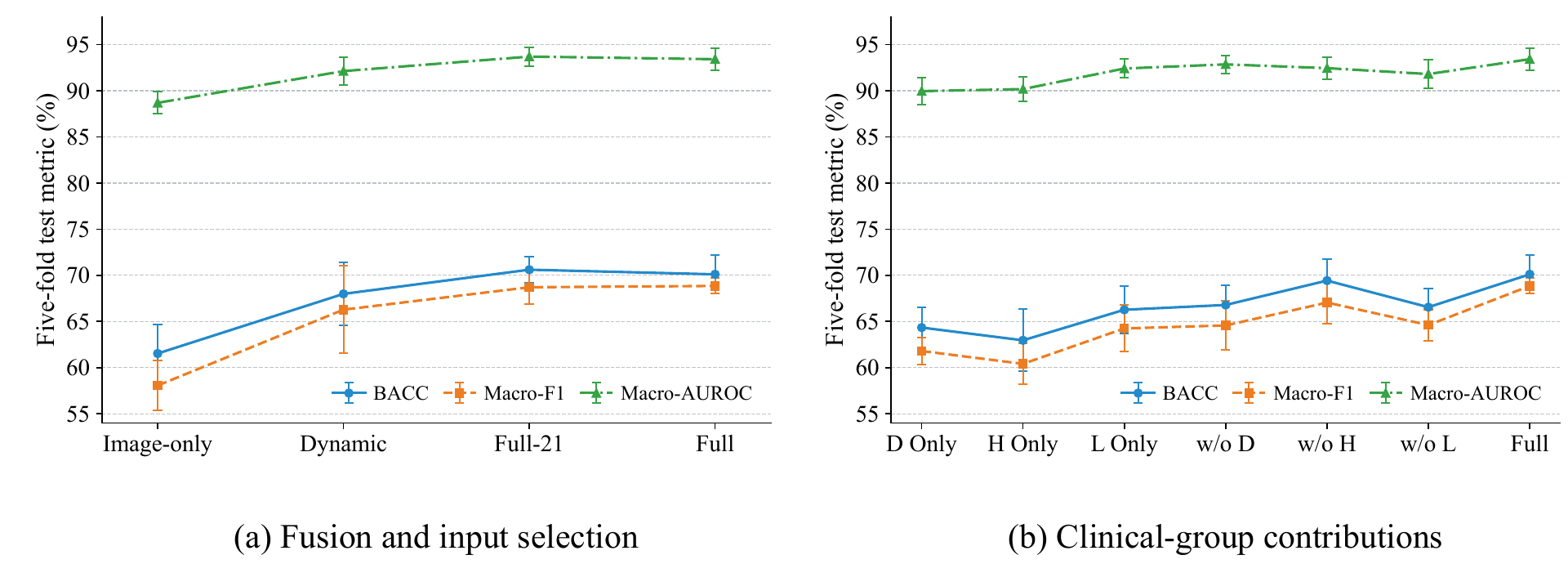}
    \caption{
    Five-fold fusion/input and clinical-group comparisons.
    Error bars denote sample standard deviations.
    Lines are visual guides across discrete configurations.
    Image-only in (a) is a reference baseline.
    }
    \label{fig:ablation}
\end{figure}

We compare fusion and input-selection alternatives and evaluate
the contributions of the three clinical groups under the same
patient-disjoint five-fold protocol; \cref{tab:ablation} reports the
retained configurations and \cref{fig:ablation} visualizes BACC,
Macro-F1, and Macro-AUROC. These comparisons assess the evaluated
configurations; they do not establish independent performance gains
for every structural component.

\begin{table}[t]
\centering
\caption{Patient-level five-fold ablation results (mean $\pm$ sample
standard deviation, \%). The full model uses the same run as the main
comparison. Bold values indicate the highest mean within each block;
the Full Model row is repeated for reference.}
\label{tab:ablation}
\setlength{\tabcolsep}{4pt}
\renewcommand{\arraystretch}{1.03}
\resizebox{\linewidth}{!}{%
\begin{tabular}{lcccc}
\toprule
Configuration & Accuracy & BACC & Macro-F1 & Macro-AUROC\\
\midrule
\multicolumn{5}{l}{\textit{Fusion and Input Selection}}\\
Dynamic Fusion & $70.04\pm4.68$ & $67.99\pm3.42$ & $66.28\pm4.74$ & $92.13\pm1.49$\\
Full-21 Fields & $72.99\pm3.53$ & $\mathbf{70.61\pm1.43}$ & $68.71\pm1.81$ & $\mathbf{93.70\pm1.02}$\\
Full Model & $\mathbf{74.17\pm2.85}$ & $70.11\pm2.11$ & $\mathbf{68.85\pm0.85}$ & $93.42\pm1.20$\\
\midrule
\multicolumn{5}{l}{\textit{Clinical Semantic Group Contributions}}\\
D Only & $65.09\pm1.75$ & $64.34\pm2.21$ & $61.80\pm1.46$ & $89.96\pm1.49$\\
H Only & $64.50\pm2.89$ & $62.96\pm3.36$ & $60.41\pm2.22$ & $90.18\pm1.29$\\
L Only & $70.80\pm0.63$ & $66.27\pm2.58$ & $64.25\pm2.51$ & $92.41\pm1.03$\\
w/o D & $70.29\pm2.92$ & $66.79\pm2.14$ & $64.57\pm2.65$ & $92.86\pm0.98$\\
w/o H & $72.53\pm2.32$ & $69.43\pm2.32$ & $67.05\pm2.27$ & $92.45\pm1.17$\\
w/o L & $68.12\pm2.45$ & $66.55\pm1.98$ & $64.61\pm1.70$ & $91.81\pm1.57$\\
Full Model & $\mathbf{74.17\pm2.85}$ & $\mathbf{70.11\pm2.11}$ & $\mathbf{68.85\pm0.85}$ & $\mathbf{93.42\pm1.20}$\\
\bottomrule
\end{tabular}}
\end{table}

\paragraph{Fusion and input selection.}
The full model exceeds Dynamic Fusion by 4.13, 2.12, 2.57, and
1.29 percentage points in Accuracy, BACC, Macro-F1, and Macro-AUROC,
respectively, so availability-normalized equal weighting achieves
higher means than the evaluated dynamic weighting variant under this
protocol. The ten-field model has higher Accuracy and Macro-F1 than
Full-21 Fields, while the latter has higher BACC and Macro-AUROC,
supporting the compact input profile as a lower-dimensional option
with similar observed performance.

\paragraph{Clinical-group contributions.}
L Only has the highest means on all four metrics among the three
single-group configurations. Removing L from the full model reduces
Accuracy, BACC, Macro-F1, and Macro-AUROC by 6.05, 3.56, 4.24,
and 1.61 points, respectively. Demographic information is also
complementary: w/o D obtains BACC of 66.79\% and Macro-F1 of
64.57\%, compared with 70.11\% and 68.85\% for the full model, and
removing H yields smaller decreases (69.43\% and 67.05\%). These
results identify L as the strongest standalone clinical group and
show that its combination with D and H achieves higher mean
performance than each single-group variant.

\paragraph{Paired uncertainty analysis.}
We retain the patient-clustered paired bootstrap analysis to examine
uncertainty in model differences. Patients are sampled with
replacement within each test fold, retaining all images of each
sampled patient. Differences are averaged equally over the five
folds. \Cref{tab:paired} reports the percentile-based 95\% confidence
intervals from 10,000 valid replicates.

\begin{table}[t]
\centering
\caption{Patient-clustered paired bootstrap comparisons. Differences
are Full Model minus the comparison model, in percentage points.
Wins count folds in which the Full Model has the higher metric.
}
\label{tab:paired}
\setlength{\tabcolsep}{3pt}
\resizebox{\linewidth}{!}{%
\begin{tabular}{lcccc}
\toprule
Comparison & $\Delta$BACC [95\% CI] & BACC wins & $\Delta$mF1 [95\% CI] & mF1 wins\\
\midrule
Full vs.\ MetaBlock & $+2.60\;[-1.52,\,6.83]$ & $4/5$ & $+4.13\;[0.69,\,7.73]$ & $5/5$\\
\bottomrule
\end{tabular}}
\end{table}

The Macro-F1 improvement over MetaBlock is positive in all five
folds and its bootstrap confidence interval is above zero, whereas
the BACC interval includes zero. The paired standardized effects are
$d_z=0.65$ for BACC and $d_z=1.35$ for Macro-F1, so the paired
comparison supports the clearest improvement over MetaBlock in
Macro-F1 under the reported evaluation protocol.

\subsection{Class-Wise and SCC-Focused Error Analysis}

\begin{figure}[t]
    \centering
    \includegraphics[width=0.90\linewidth]{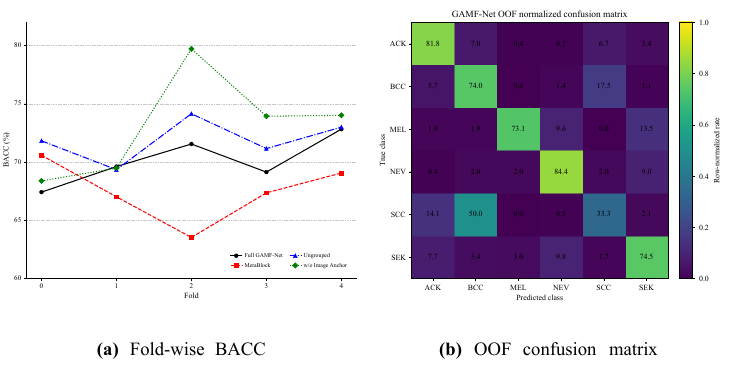}
    \caption{
    Fold-wise and class-wise analysis.
    (a) BACC across the five patient-disjoint test folds.
    (b) Row-normalized confusion matrix from the aggregated
    out-of-fold predictions of \method{}.
    }
    \label{fig:diagnostic-analysis}
\end{figure}

\begin{table}[t]
\centering
\caption{
Class-wise five-fold results
(mean $\pm$ sample standard deviation, \%).
}
\label{tab:perclass}
\setlength{\tabcolsep}{7pt}
\begin{tabular}{lcc}
\toprule
Class & Recall & AUROC\\
\midrule
ACK & $81.89\pm3.83$ & $95.50\pm1.30$\\
BCC & $74.13\pm4.46$ & $92.86\pm1.27$\\
MEL & $72.61\pm9.85$ & $98.87\pm0.84$\\
NEV & $84.22\pm3.22$ & $97.95\pm0.59$\\
SCC & $33.22\pm4.47$ & $79.95\pm5.69$\\
SEK & $74.62\pm4.90$ & $95.39\pm2.70$\\
\bottomrule
\end{tabular}
\end{table}

NEV and ACK achieve the highest recalls ($84.22\%$ and $81.89\%$);
MEL obtains the highest mean AUROC ($98.87\%$) but with substantial
fold-wise recall variation, consistent with its small sample size.

SCC remains the principal failure class: in the aggregated
out-of-fold predictions, \method{} obtains an SCC precision of
$23.70\%$, recall of $33.33\%$, F1 of $27.71\%$, AUROC of
$78.81\%$, and AUPRC of $24.20\%$; the pooled recall differs
slightly from the $33.22\pm4.47\%$ in \cref{tab:perclass} because
the former merges all five test folds, whereas the latter averages
fold-level recalls.

Relative to the image-only model, \method{} increases SCC AUROC
from $73.31\%$ to $78.81\%$ and AUPRC from $21.55\%$ to
$24.20\%$, but does not improve SCC precision, recall, or F1 under
the default argmax decision rule. MetaBlock remains stronger for
SCC-specific recognition (recall of $42.71\%$, F1 of $34.89\%$, and
AUROC of $83.34\%$). Thus, the overall Macro-F1 improvement of
\method{} does not represent a uniform gain across all classes.

Of the 192 SCC samples, 64 are correctly classified; of the 128
errors, 96 cases ($75.00\%$) are predicted as BCC and 27 cases
($21.09\%$) as ACK, and the confusion is partly reciprocal, with 148
BCC samples assigned to SCC. The main class-level limitation is
therefore the decision boundary between SCC and BCC rather than
uniformly poor ranking across all lesion categories.

\subsection{Discussion and Limitations}

The experiments support four main observations. First, structured
clinical information substantially improves overall and class-balanced
recognition over the image-only baseline. Second, the constrained
availability-normalized fusion outperforms the evaluated dynamic
fusion variant. Third, the compact ten-field profile performs
comparably to the full 21-field input, and the lesion-presentation
group provides the strongest clinical complement. Semantic grouping
supports clinical organization and group-level interpretation, while
the image anchor preserves a separate visual pathway and class-level
residual corrections; the ablations characterize the evaluated
configurations rather than isolate each component.

This study has several limitations. PAD-UFES-20 was collected from a
specific population and acquisition setting, so generalization across
devices, regions, and institutions remains unverified, and the
availability-aware formulation has been evaluated only under the
naturally occurring missingness distribution rather than under
progressively increasing field absence or systematic removal of
entire clinical groups. In addition, SCC recognition remains
substantially weaker than recognition of the other classes, primarily
because of SCC--BCC confusion. Finally, the patient-clustered
bootstrap quantifies uncertainty due to test-patient sampling but not
variability across repeated training runs; external validation and
repeated-training analysis would be required for stronger conclusions
about generalization and optimization stability.

\section{Conclusion}
We proposed \method{}, which combines smartphone skin lesion images
and structured clinical information through semantic grouping,
field-availability-aware encoding, and image-anchored class-level
residual fusion.
Under a patient- and lesion-disjoint five-fold protocol, \method{}
achieves the highest mean Accuracy, BACC, Macro-F1, and
Macro-AUROC under the unified protocol, with the patient-clustered
analysis providing the strongest evidence for a consistent Macro-F1
improvement over MetaBlock. Constrained equal-weight fusion is more
effective than the evaluated dynamic fusion variant, while the
compact ten-field profile retains performance comparable to the full
21-field input.
Future work will focus on SCC--BCC discrimination, structured
missing-data scenarios, and validation across backbones, devices,
and institutions.

\begin{credits}
\subsubsection{\ackname} This work was supported by
the National Natural Science Foundation of China (No. 62001184),
the Guangdong Basic and Applied Basic Research Foundation (No.
2025A1515110445), the Projects of Educational Commission of Guangdong
Province (Nos. 2025ZDZX1038, 2025KTSCX49, 2023ZDZX1025),
the Huizhou University Independent Innovation Capability Enhancement
Plan (No. HZU202535), and the College Students' Innovative
Entrepreneurial Training Plan Program (Nos. 202410577008,
202610577034).
\end{credits}

\bibliographystyle{splncs04}
\bibliography{references}

\end{document}